\documentclass[sn-mathphys]{sn-jnl}

\usepackage{ulem} 

\jyear{2021}%

\theoremstyle{thmstyleone}%
\theoremstyle{thmstyletwo}%

\theoremstyle{thmstylethree}%

\begin{document}

\title[Reconstruction of Incomplete Wildfire Data using Deep Generative Models]{Reconstruction of Incomplete Wildfire Data using Deep Generative Models}

\author[1]{\fnm{Tomislav} \sur{Ivek}}\email{tivek@ifs.hr}
\equalcont{These authors contributed equally to this work.}

\author*[2]{\fnm{Domagoj} \sur{Vlah}}\email{domagoj.vlah@fer.hr}
\equalcont{These authors contributed equally to this work.}

\affil[1]{\orgname{Institut za fiziku}, \orgaddress{\street{Bijenička 46}, \city{Zagreb}, \postcode{HR-10000}, \country{Croatia}}}

\affil*[2]{\orgdiv{Department of Applied Mathematics}, \orgname{University of Zagreb, Faculty of Electrical Engineering and Computing}, \orgaddress{\street{Unska 3}, \city{Zagreb}, \postcode{HR-10000}, \country{Croatia}}}

\abstract{We present our submission to the Extreme Value Analysis 2021 Data Challenge in which teams were asked to accurately predict distributions of wildfire frequency and size within spatio-temporal regions of missing data. For the purpose of this competition we developed a variant of the powerful variational autoencoder models dubbed the Conditional Missing data Importance-Weighted Autoencoder (CMIWAE). Our deep latent variable generative model requires little to no feature engineering and does not necessarily rely on the specifics of scoring in the Data Challenge. It is fully trained on incomplete data, with the single objective to maximize log-likelihood of the observed wildfire information. We mitigate the effects of the relatively low number of training samples by stochastic sampling from a variational latent variable distribution, as well as by ensembling a set of CMIWAE models trained and validated on different splits of the provided data. The presented approach is not domain-specific and is amenable to application in other missing data recovery tasks with tabular or image-like information conditioned on auxiliary information. }




\maketitle

\section{Introduction}\label{sec1}
Wildfires are unplanned and often uncontrolled fires of combustible vegetation. They depend on various natural or man-made conditions such as weather, the vegetation type and cover, animal populations, terrain, human activity etc., and may stay localized or spread across the area of some susceptible geographical region. A direct danger to human lives as well as an environmental and economic hazard, the extreme properties and consequences of wildfires are the subject of continued attempts at modeling using various statistical methods, see eg.\ \cite{Preisler04,Xi19,Pereira19}.

The EVA 2021 Data Competition \cite{Opitz22} presents a spatio-temporal dataset of USA wildfire activity in the United States during a 23-year period arranged on a spatial grid. Wildfire activity itself is characterized by two variables, the aggregated burnt area (BA) and the wildfire count (CNT) for each spatial grid cell in a span of one month. Additionally, it is accompanied by 35 auxiliary variables describing the land cover, meteorological, and geological conditions for each spatio-temporal cell. About 14.2\% of both BA and CNT values have been masked and replaced by a tag denoting missing data. The missingness mask covers even years only and is spatio-temporally not uniformly random, but forms clusters in both space and time.

The competition task is to estimate the cumulative distribution functions for the number of wildfires and the aggregated burnt area in the cells with missing data. Sets of 28 roughly exponential severity thresholds are given, $\mathcal{U}_\mathrm{CNT}$, and $\mathcal{U}_\mathrm{BA}$, on which the probabilities $\hat{p}_{\mathrm{CNT},i}(u) = \mathbb{P}(\mathrm{CNT}_i\leq u),\ u\in\mathcal{U}_\mathrm{CNT}$ and $\hat{p}_{\mathrm{BA},i}(u) = \mathbb{P}(\mathrm{BA}_i\leq u),\ u\in\mathcal{U}_\mathrm{BA}$ need to be evaluated for all missing cells $i$.

The quality of predictions is scored (smaller is better) by the competition organizer for BA and CNT separately as the following sums over missing data:
\begin{align}
S_\mathrm{CNT}&=\sum_{i=1}^{k_\mathrm{CNT}} \sum_{u\in\mathcal{U}_\mathrm{CNT}}\omega_\mathrm{CNT}(u)\cdot \left(\mathbb{I}\left(u\geq\mathrm{CNT}_i\right)-\hat{p}_{\mathrm{CNT},i}(u)\right)^2,\\
S_\mathrm{BA}&=\sum_{i=1}^{k_\mathrm{BA}} \sum_{u\in\mathcal{U}_\mathrm{BA}}\omega_\mathrm{CNT}(u)\cdot \left(\mathbb{I}\left(u\geq\mathrm{BA}_i\right)-\hat{p}_{\mathrm{BA},i}(u)\right)^2,
\label{eq:score}
\end{align}
where $\mathbb{I}$ is the standard indicator function $\mathbb{I}(u<x)=1$ if $u\leq x$ and 0 otherwise. The weight functions $\omega$ award good predictions in the extremes: $\omega_\mathrm{CNT}(u)=\hat{\omega}_\mathrm{CNT}(u)/\hat{\omega}_\mathrm{CNT}(u_{28})$, $\hat{\omega}_\mathrm{CNT}(u)=1 - (1 + (u+1)^2/1000)^{-1/4}$, and  
$\omega_\mathrm{BA}(u)=\hat{\omega}_\mathrm{BA}(u)/\hat{\omega}_\mathrm{BA}(u_{28})$, $\hat{\omega}_\mathrm{BA}(u)=1 - (1 + (u+1)/1000)^{-1/4}$. Predictions are finally ranked three times based on $S_\mathrm{CNT}$, $S_\mathrm{BA}$, and $S_\textrm{total}=S_\mathrm{CNT}+S_\mathrm{BA}$.\cite{Opitz22}

In the wildfire dataset, each of 161 data slices along the temporal axis can be regarded as an image with multiple features per spatial cell. Among deep learning models, architectures based on learnable convolutions, so-called convolutional neural networks, are extensively used in image data refinement, classification, restoration, and generation with great success. Among many available references see eg. \cite{Schlemper2017_2,stylegan}. Most are fitted on a number of training images to reduce them to their most salient features and by doing so create their latent space, which is most suitable for the task at hand (classification, denoising, inpainting, etc.).  Typically, a model would be trained on a complete dataset to generate an output close to the associated input according to a chosen metric. Then, missing data could be recovered by taking the trained autoencoder's output of the input with missing data. This methodology requires training data to be complete, with no missing information, and datasets quite large, which are both obvious drawbacks.

Variational autoencoder (VAE) \cite{Kingma14} and its more expressive successor, the importance-weighted autoencoder (IWAE) \cite{Burda16}, seem to be more appropriate for smaller datasets. As opposed to conventional neural network models, which create one-to-one mappings between the input and its latent representation, VAE and IWAE map the input to its own distribution in the latent space from which latent vectors are sampled and mapped to the desired output. This sampling inside the model evaluation creates robust latent representations, while using small datasets. The training procedure optimizes the log-likelihood of the output probability distribution with respect to input data. Unfortunately, even though a trained model can be used to recover missing data, the training dataset is still required to be fully complete.

Mattei et al.\ recently developed the missing data importance-weighted autoencoder (MIWAE) \cite{Mattei19} in an effort to repair missing data using models trained on the incomplete input. A MIWAE model is fitted by maximizing the log-likelihood on observed data only, and is suitable for smaller datasets where the pattern of missingness does not depend on missing data itself. This model has subsequently been used as a foundation for more specialized models, such as the not-MIWAE model \cite{Ipsen21}, and its performance may have already been superseded on certain datasets by other approaches such as the miss-IWEM \cite{Kim20}.

In the following text we give a short overview of the MIWAE approach. As {\it Team BlackBox}, we then expand on a MIWAE by conditioning the modeled probabilities on image-like auxiliary features, which are in majority, always observed, and thus do not need to be modeled. The resulting new model, the conditional missing data importance-weighted autoencoder (CMIWAE), accurately predicts missing features of the USA wildfire dataset, all with little to no feature engineering and after being trained on a single optimization goal of maximizing the log-likelihood of observed data.

\section{Bayesian inference of missing data using autoencoders}

In general terms, data may be missing completely at random (MCAR, the probability of being missing does not depend on data), missing at random (MAR, the probability of being missing may be conditioned on observed data only), or missing not-at-random (MNAR, the probability of being missing may be conditioned on both observed and missing data). For the task of predicting frequency and size of wildfires within spatially and temporally correlated clusters of missing data, a simple MCAR assumption may be unjustified. On the other hand, the MNAR case is tractable in theory, eg. see the not-MIWAE model \cite{Ipsen21}, but it requires significant effort to model the probability distribution of missingness and the training procedure is computationally intensive. In this work we take the USA wildfire data as missing at random and build a model on this assumption.

\subsection{Missing Data Importance-weighted Autoencoder}\label{sec:MIWAE}

Autoencoders are models fitted to best approximate the identity function by mapping it to and then from latent space, in effect preserving only the most relevant features of the original data. They are commonly robust to input noise and, if fitted adequately, may generalize to data not seen during the fitting procedure. The Missing Data Importance-weighted Autoencoder \cite{Mattei19} assumes independent and identically distributed (i.i.d.) samples of observable data $(\mathbf{x}_1, ..., \mathbf{x}_N)$, $\mathbf{x}_i\in\mathcal{K}$, with $p$ different features present in the data. If some features are missing (not observed), each sample $i \in \{1, ..., n\}$ can be split into the observed and missing features $\mathbf{x}_i^\mathrm{o}$ and $\mathbf{x}_i^\mathrm{m}$, respectively. The indices of observed features can then be represented by binary vectors $\mathbf{m}_i \in \{0, 1\}^p$ where $m_{ij}=1$ if feature $j$ of sample $i$ is observed, and $m_{ij}=0$ if feature $j$ of sample $i$ is missing.

Latent variable models attempt to describe this high-dimensional observable data as a function of some so-called latent variables $(z_1, ..., z_n), z_i \in \mathbb{R}^d$ where typically $d<p$. Deep latent variable generative models \cite{Kingma14,Rezende14} generally assume that $(\mathbf{x}_i, \mathbf{z}_i)_{i\leq n}$ are driven by the generative model $\mathbf{z}\sim p(\mathbf{z}),\ p_{\mathbf{\theta}}(\mathbf{x}\vert\mathbf{z})=\mathrm{\varPhi}\left(\mathbf{x}\vert f_{\theta}(\mathbf{z})\right)$. Here, $p(\mathbf{z})$ is the prior distribution of the latent variables which may have fittable parameter but is typically taken to be a standard Gaussian distribution.  Further, $\left(\mathbf{\varPhi}(\cdot\vert\mathbf{\boldsymbol{\eta}})\right)_{\mathbf{\boldsymbol{\eta}}\in H}$ is the parametric family of distributions over the observable data called the observation model, and $f_{\boldsymbol{\theta}}:\mathbb{R}^{d}\longrightarrow H$ is the so-called decoder function which is typically a neural network with parameters $\mathbf{\theta}\in\mathbf{\Theta}$.

A deep latent variable model is usually trained to maximize a lower bound of the log-likelihood function. Under the missing-at-random assumption (missingness probability may depend on the observed data but not on the missing data), in the absence of complete data, we would like the trained model to maximize the log-likelihood of all of the observed data $\mathbf{x}_{1}^{\mathrm{o}},\ldots,\mathbf{x}_{n}^{\mathrm{o}}$
\begin{equation}
\ell(\boldsymbol{\theta})=\sum_{i=1}^{n}\log p_{\mathbf{\theta}}(\mathbf{x}_{i}^{\mathrm{o}})=\sum_{i=1}^{n}\log\int p_{\mathbf{\theta}}(\mathbf{x}_{i}^{\mathrm{o}}\vert\mathbf{z})p(\mathbf{z})d\mathbf{z}.
\label{eq:loglikelihood}
\end{equation}
Note that the mask is not part of model input, rather it serves to define the optimization objective by differentiating between observed and missing data.

Such an integral is unfortunately most often intractable. The authors of MIWAE derive a lower bound of $\ell(\mathbf{\theta})$ which is easier to maximize. For this purpose, amortised variational inference is introduced \cite{Gershman14,Kingma14,Rezende14} by taking a variational distribution  $q_{\boldsymbol{\gamma}}(\mathbf{z}\vert\mathbf{x}^{\mathrm{o}})$ which takes place of the intractable posterior distribution $p_{\mathbf{\theta}}(\mathbf{z}\vert\mathbf{x}^{\mathrm{o}})$. This variational distribution is defined as $q_{\boldsymbol{\gamma}}(\mathbf{z}\vert\mathbf{x}^{\mathrm{o}})=\mathrm{\varPsi}\left(\mathbf{z}\vert g_{\boldsymbol{\gamma}}\left(\iota(\mathbf{x}^{\mathrm{o}})\right)\right)$. In this expression, $\iota$ is a simple imputation function chosen beforehand which maps $\mathbf{x}^\mathrm{o}$ into a complete input vector $\iota(\mathbf{x}^{\mathrm{o}})\in\mathcal{X}$, $\iota(\mathbf{x}^{\mathrm{o}})^\mathrm{o}=\mathbf{x}^{\mathrm{o}}$, and $\left(\mathbf{\varPsi}(\cdot\vert\mathbf{\boldsymbol{\kappa}})\right)_{\mathbf{\boldsymbol{\kappa}}\in\mathcal{K}}$ is the variational family of simple distributions over $\mathbb{R}^{d}$. The last remaining piece of amortised variational inference is the encoder function $g_{\boldsymbol{\boldsymbol{\gamma}}}:\mathcal{X}\longrightarrow\mathcal{K}$ which is typically a neural network with parameters $\mathbf{\gamma}\in\mathbf{\Gamma}$ which serves to map each data point into the parameters of the variational family $\mathbf{\varPsi}$.

The MIWAE stochastic lower bound is then obtained as the importance-sampled Monte Carlo approximation of Eq.\ \ref{eq:loglikelihood} with proposal distributions $q_{\boldsymbol{\gamma}}(\mathbf{z}\vert\mathbf{x}_{i}^{\mathrm{o}})$ and $K\in\mathbb{N}^\ast$ samples, 
\begin{equation}
\mathcal{L}_{K}(\boldsymbol{\theta},\boldsymbol{\gamma})=\sum_{i=1}^{n}\mathbb{E}_{\mathbf{z}_{i1},\ldots,\mathbf{z}_{iK}\sim q_{\boldsymbol{\gamma}}(\mathbf{z}\vert\mathbf{x}^{\mathrm{o}})}\left[\log\frac{1}{K}\sum_{k=1}^{K}\frac{p_{\mathbf{\theta}}(\mathbf{x}_{i}^{\mathrm{o}}\vert\mathbf{z}_{ik})p(\mathbf{z}_{ik})}{q_{\boldsymbol{\gamma}}(\mathbf{z}_{ik}\vert\mathbf{x}_{i}^{\mathrm{o}})}\right].
\label{eq:MIWAE_bound}
\end{equation}
$K=1$ recovers the well-known evidence lower bound typically used by variational autoencoders. The MIWAE bound becomes tighter as $K$ grows larger, $\mathcal{L}_{1}(\boldsymbol{\theta},\boldsymbol{\gamma})\leq\mathcal{L}_{2}(\boldsymbol{\theta},\boldsymbol{\gamma})\leq\ldots\leq\mathcal{L}_{K}(\boldsymbol{\theta},\boldsymbol{\gamma})\xrightarrow[K\longrightarrow\infty]{}\ell(\boldsymbol{\theta})$. Moreover, under some mild moments conditions on the importance weights, \cite{Domke18} the MIWAE lower bound has the useful property
\begin{equation}
\mathcal{L}_{K}(\boldsymbol{\theta},\boldsymbol{\gamma})=_{K\longrightarrow\infty}\ell(\boldsymbol{\theta})+O(1/K).
\label{eq:bound_limit}
\end{equation}
This result shows that a large $K$ will lead to optimizing a tight bound of the log-likelihood even if the imputation function $\iota$ gives poor results on its own. This allows us to apply simple and scalable imputation on model input while training the model, such as zero imputation, and still expect good performance from the trained model.

\subsection{Reparameterization trick}\label{sec:reparameter}
Objective Eq.\ \ref{eq:MIWAE_bound} is typically optimized using a variation of gradient descent algorithm which finds $\boldsymbol{\theta},\boldsymbol{\gamma}$ that minimize $-\mathcal{L}_{K}(\boldsymbol{\theta},\boldsymbol{\gamma})$. In order to do that, $\mathbf{z}_{ik}$ sampled from the variational distribution $q_{\boldsymbol{\gamma}}(\mathbf{z}\vert\mathbf{x}_{i}^{\mathrm{o}})$ must be differentiable over parameters $\boldsymbol{\gamma}$. The reparameterization trick is a way to rewrite the stochastic variable so that the stochastic sampling is independent of the parameters $\boldsymbol{\gamma}$ with respect to which we take the gradient.\cite{Williams1992} 

Typically it is assumed that $q_{\boldsymbol{\gamma}}(\mathbf{z}\vert\mathbf{x}_{i}^{\mathrm{o}})$ is a multivariate Gaussian distribution $N(\boldsymbol{\mu}_i, \boldsymbol{\sigma}_i)$ with mean $\boldsymbol{\mu}_i\in\mathbb{R}^d$ and diagonal standard deviation $\boldsymbol{\sigma}_i\in\mathbb{R}^d$. The samples $\boldsymbol{z}_{ik}$ can be reparameterized as $\boldsymbol{z}_{ik}=\boldsymbol{\mu}_i+\boldsymbol{\sigma}_i\boldsymbol{\varepsilon}_k$ where the random variable $\boldsymbol{\varepsilon}\sim N(0,1)^d$ is independent of distribution parameters.

\subsection{Conditioning the model on auxiliary data}\label{sec:CMIWAE}
The MIWAE framework is quite useful when data might be missing from each of $p$ features in any data sample $\mathbf{x}\in\mathcal{K}$. However, if certain features are guaranteed to always be observed, it is unnecessary to model them for imputation. Such is the case eg.\ of the USA wildfire data: burnt area and wildfire count features may be missing, but all the other provided features (meteorological, geological, geographical, chronological etc.) are complete for each sample.

It is reasonable to separate the always-complete data features into auxiliary data $\mathbf{c}\in\mathcal{C}$ and keep the data features which might be missing in $\mathbf{x}\in\mathcal{K}$. We can straightforwardly rewrite the optimization objective Eq.\ \ref{eq:loglikelihood} to condition it on $\mathbf{c}$:
\begin{equation}
\ell(\boldsymbol{\theta})=\sum_{i=1}^{n}\log p_{\mathbf{\theta}}(\mathbf{x}_{i}^{\mathrm{o}}\vert\mathbf{c}_{i})=\sum_{i=1}^{n}\log\int p_{\mathbf{\theta}}(\mathbf{x}_{i}^{\mathrm{o}}\vert\mathbf{z},\mathbf{c}_{i})p_{\boldsymbol{\alpha}}(\mathbf{z}\vert\mathbf{c}_{i})d\mathbf{z}.
\label{eq:CMIWAE_loglikelihood}
\end{equation}
Note that the artificial (but still useful) split of observed features into $\mathbf{x}^\mathrm{o}$ and  $\mathbf{c}$ means Eq.\ \ref{eq:CMIWAE_loglikelihood} is akin to marginalizing the likelihood inside the logarithm of Eq.\ \ref{eq:loglikelihood} over the features $\mathbf{c}$. Next, we assume the variational distribution takes the form of $q_{\boldsymbol{\gamma}}(\mathbf{z}\vert\mathbf{x}^{\mathrm{o}},\mathbf{c})=\mathrm{\varPsi}\left(\mathbf{z}\vert g_{\boldsymbol{\gamma}}\left(\iota(\mathbf{x}^{\mathrm{o}}),\mathbf{c}\right)\right)$, and the observation model $p_{\mathbf{\theta}}(\mathbf{x}\vert\mathbf{z},\mathbf{c})=\mathrm{\varPhi}\left(\mathbf{x}\vert f_{\theta}(\mathbf{z},\mathbf{c})\right)$ where the neural networks $g_{\boldsymbol{\gamma}}$ and $f_{\boldsymbol{\theta}}$ now also depend on $\mathbf{c}$. As a departure from MIWAE, which conventionally uses a standard Gaussian prior, the prior is now parameterized with $\boldsymbol{\alpha}$ as $p_{\boldsymbol{\alpha}}(\mathbf{z}\vert\mathbf{c}_{i})=\mathrm{\Xi}\left(\mathbf{z}\vert h_{\alpha}(\mathbf{c})\right)$. $\left(\Xi(\cdot\vert\mathbf{\boldsymbol{\xi}})\right)_{\mathbf{\boldsymbol{\xi}}\in H^\prime}$ is the parametric family of distributions over the latent space, and $h_{\boldsymbol{\alpha}}:\mathcal{C}\longrightarrow H^\prime$ is the auxiliary encoder function which is typically a neural network with parameters $\boldsymbol{\alpha}\in\mathbf{A}$. In this work, we use a fittable Gaussian prior, but other reparameterizable distribution families can also be used.

Analogously to the MIWAE bound, we obtain the conditional MIWAE (CMIWAE) bound
\begin{equation}
\mathcal{L}_{K}(\boldsymbol{\theta},\boldsymbol{\gamma})=\sum_{i=1}^{n}\mathbb{E}_{\mathbf{z}_{i1},\ldots,\mathbf{z}_{iK}\sim q_{\boldsymbol{\gamma}}(\mathbf{z}\vert\mathbf{x}^{\mathrm{o}},\mathbf{c}_{i})}
\left[\log\frac{1}{K}\sum_{k=1}^{K}w_{ik}\right],
\label{eq:CMIWAE_bound}
\end{equation}
where the weights $w_{ik}$ are given by
\begin{equation}
w_{ik} = \frac{p_{\mathbf{\theta}}(\mathbf{x}_{i}^{\mathrm{o}}\vert\mathbf{z}_{ik},\mathbf{c}_{i})p_{\boldsymbol{\alpha}}(\mathbf{z}_{ik}\vert\mathbf{c}_{i})}{q_{\boldsymbol{\gamma}}(\mathbf{z}_{ik}\vert\mathbf{x}_{i}^{\mathrm{o}},\mathbf{c}_{i})}.
\label{eq:nnweights}
\end{equation}
The above optimization objective is well-suited to model the missing-at-random wildfire features while conditioning the observation model, prior, as well as the variational distribution on the auxiliary data features which are always complete.

\subsection{Double-reparameterized gradient estimation}\label{sec:DReG}
The derivative of the optimization objective Eq.\  \ref{eq:MIWAE_bound} over model parameters often results in gradients with an overly large variance which destabilize most common optimization algorithms.\cite{Roeder17,Rainforth18} Our own observations mirror these conclusions. To obtain faster numerical convergence of model training, we have implemented the double-reparameterized gradient estimation (DReG) \cite{Tucker19} which obtains an unbiased, low-variance estimate of the gradient over $\boldsymbol{\gamma}$
\begin{multline}
\nabla{}_{\boldsymbol{\gamma}}  \mathbb{E}_{\mathbf{z}_{i1},\ldots,\mathbf{z}_{iK}\sim q_{\boldsymbol{\gamma}}(\mathbf{z}\vert\mathbf{x}^{\mathrm{o}},\mathbf{c}_{i})}\left[\log\left(\frac{1}{K}\sum_{k=1}^K w_{ik}\right)\right] = \\
= \mathbb{E}_{\mathbf{\varepsilon}_{i1},\ldots,\mathbf{\varepsilon}_{iK}\sim N(0,1)}\left[ \sum_{k=1}^K \left(\frac{w_{ik}}{\sum_j w_{jk}}\right)^2 \frac{\partial \log w_{ik}}{\partial \mathbf{z}_{ik}} \frac{\partial \mathbf{z}_{ik}}{\partial \boldsymbol{\gamma}}\right].
\end{multline}
(Note: If we were to follow the conventional model naming scheme, ours would more precisely be called CMIWAE-DReG which seems a bit unwieldy.)

\subsection{Recovering the cumulative distribution of missing data}\label{sec:cum_distro}
A trained MIWAE-type model can be used to efficiently estimate missing data. Let $y(\mathbf{x}^\mathrm{m})$ be a desired integrable function of the missing data. Then, just as for MIWAE \cite{Mattei19}, a CMIWAE estimate can be derived using self-normalized importance sampling:
\begin{equation}
\mathbb{E}\left[y\left(\mathbf{x}^\mathrm{m}\right)\vert\mathbf{x}^\mathrm{o},\mathbf{c}\right] = \int y(\mathbf{x}^\mathrm{m}) p_{\boldsymbol{\theta}}(\mathbf{x}^\mathrm{m}\vert\mathbf{x}^\mathrm{o},\mathbf{c})d\mathbf{x}^\mathrm{m}
\approx \sum_j \overline{w}_j y\left(\mathbf{x}_{(j)}^\mathrm{m}\right),
\label{eq:general_quantity_estimate}
\end{equation}
where $(\mathbf{x}_{(j)}^\mathrm{m},\ \mathbf{z}_{(j)})$ are i.i.d. samples obtained by ancestral sampling from $p_{\boldsymbol{\theta}}(\mathbf{x}^\mathrm{m}\vert\mathbf{z},\mathbf{x}^\mathrm{o},\mathbf{c}) q_{\boldsymbol{\gamma}}(\mathbf{z}\vert\mathbf{x}^{\mathrm{o}},\mathbf{c})$, and $\overline{w}_j = w_j / \sum_k w_k$ are normalized weights $w_j$ given by Eq.\ \ref{eq:nnweights}.

In this work we are primarily interested in estimating the cumulative distribution function of the missing data $P_{\boldsymbol{\theta}}(\mathbf{x}^\mathrm{m}\vert\mathbf{x}^\mathrm{o},\mathbf{c})$. We can skip one level of ancestral sampling in Eq.\ \ref{eq:general_quantity_estimate} by observing that the probability density distribution $p_{\boldsymbol{\theta}}(\mathbf{x}^\mathrm{m}\vert\mathbf{z}_{j},\mathbf{x}^\mathrm{o},\mathbf{c})$ is directly provided by the model for each sample $\mathbf{z}_{j}$. Hence, instead of Eq.\ \ref{eq:general_quantity_estimate} it is more efficient to use a Rao-Blackwell-Kolmogorov-type estimator \cite{Ipsen21},
\begin{equation}
\mathbb{E}\left[p_{\boldsymbol{\theta}}(\mathbf{x}^\mathrm{m}\vert\mathbf{x}^\mathrm{o},\mathbf{c})\right] \approx \sum_j \overline{w}_j p_{\boldsymbol{\theta}}(\mathbf{x}^\mathrm{m}\vert\mathbf{z}_{j},\mathbf{x}^\mathrm{o},\mathbf{c}).
\end{equation}
Integrating this expression we finally obtain an estimate for the cumulative distribution function of the missing data,
\begin{equation}
\mathbb{E}\left[P_{\boldsymbol{\theta}}(\mathbf{x}^\mathrm{m}\vert\mathbf{x}^\mathrm{o},\mathbf{c})\right] \approx \sum_j \overline{w}_j \int p_{\boldsymbol{\theta}}(\mathbf{x}^\mathrm{m}\vert\mathbf{z}_{j},\mathbf{x}^\mathrm{o},\mathbf{c})d\mathbf{x}^\mathrm{m}.
\end{equation}

\section{Applying CMIWAE to the problem of missing wildfire data}

Our model, training and evaluation code is available at \url{https://github.com/Blackbox-EVA2021/CMIWAE}.

The data are available from the organizers of the Extreme Value Analysis 2021 Data Challenge, \url{https://www.maths.ed.ac.uk/school-of-mathematics/eva-2021/competitions/data-challenge}. Data are also available from the authors upon reasonable request and with permission of the organisers of the Data Challenge.

\subsection{Data preparation and dimensionality}\label{sec:data_prep}
One significant advantage of deep neural models is only a minimal need for feature engineering. In our case this amounts to data normalization, calculating logarithms of variables which span many orders of magnitude, and arranging the values into a tensor.

We transformed the originally provided USA wildfire dataset into a masked four-dimensional real tensor of shape $T \times d_{in} \times W \times H$, where $T=161$, $d_{in}=41$, $W=128$, and $H=64$. The first dimension is time, as the total number of months in dataset equals to $23$ years $\times$ $7$ months per year. The second dimension corresponds to $d_{in}=2+2+37$ spatiotemporal data features. The third and fourth dimensions are the spatial $x$ and $y$ grid cell coordinates, slightly expanded by padding on the edges, for performance reasons to be of the form $2^n$.

Further, three missingness masks are provided. The first mask is of shape $W \times H$ for all the valid continental USA geographical locations on a $0.5^{\circ} \times 0.5^{\circ}$ grid. The other two are $T\times W \times H$-shaped masks for grid positions of the known values of CNT and BA.

Out of $41=2+2+37$ features per grid cell in the prepared dataset, the first 2 are CNT and BA and the last $37$ are obtained from the original $35$ auxiliary features. Here, missing values of CNT and BA are consistently imputed with value $0$ according to their respective masks. Additionally, 2 channels of $\ln(\mathrm{CNT}+1)$ and $\ln(\mathrm{BA}+1)$ are appended to help the model reason about logarithmic scales inherent to CNT and BA. As for the auxiliary data in the remaining 37 features per grid cell, the original year plus month pair is transformed to a continuous real time variable, allowing the model to potentially uncover eg.\ any long-term climate trends. Original month data is embedded into a fittable 3-dimensional real vector to account for learnable seasonality. Longitude, latitude, height related data, the meteorological variables, and the continuous time are normalized by subtracting their mean and dividing by standard deviation over all individual spatiotemporal locations.

Hence, the data $\mathbf{x}$ that might be missing can be regarded as a tensor of shape $T \times 4 \times W \times H$, and the auxiliary data $\mathbf{c}$ as a tensor of shape $T \times 37 \times W \times H$. Samples are taken from the dataset by slicing the data tensor along the temporal dimension, for each sample $i$ resulting in tensors $\mathbf{x}_i$, $\mathbf{c}_i$ of shapes $1 \times 4 \times W \times H$ and $1 \times 37 \times W \times H$, respectively, which are used as the model input. Each sample is accompanied by the missingness mask $\mathbf{m}_i$ of shape $1 \times 2 \times W \times H$, which is never passed on to the model but exclusively used by the optimization objective to indicate grid cells over which to sum the log-likelihoods.

\subsection{Training and validation datasets and their role in training}

We separated the provided data, per year basis, in two disjoint subsets, the so-called (conventional in machine learning) the training and the validation dataset. The training dataset is used for the purpose of objective optimization by fitting model parameters. The validation dataset is not used in training directly, but is used to measure performance of the model on unseen data during training, by continuous assessing of the training convergence. The test dataset consists of the missing data that was withheld from the contestants, which is the masked CNT and BA data in even dataset years. The test dataset was not available to us during the competition phase and so could not be used for model training or optimization of model hyperparameters. Only later, after the competitions final ranking was determined, the test dataset was provided by the competition organizers, so we could use it to compute scores of different variants of our model, as described in Section \ref{sec:results}. Notice that we differ in terminology from the competition problem description in \cite{Opitz22}. There, the training dataset refer to what is the union of training and validation dataset in this paper. Also, there the validation dataset refer to what we call the test dataset.

Out of $23$ years of the given data, we decided to use only $2$ years for the validation dataset and the rest we used for the training dataset. This gives us a total of ${23 \choose 2} = 253$ combinations of dataset splits to chose from. The decision to use yearly, and not monthly, train-validation split was motivated by the need to avoid information bleed between the train and validation datasets, as would presumably happen if neighbouring months were placed into separate datasets.

During training, the value being optimized is the estimated log-likelihood lower bound from Eq.\ \ref{eq:CMIWAE_bound}. For this competition, at the end of a single training procedure we chose to keep the model with the lowest estimated total score (which may not be the model with the highest log-likelihood). We estimate the total score using the validation data in the following way. Missingness masks from the non-validation BA and CNT samples are randomly selected and applied to the validation dataset. The validation data with additional missingness is used as input to the model, and the model reconstructs the missing data, including the newly removed values. Score is calculated at the end of each gradient update (training epoch) based on the values of known BA and CNT which were removed and finally normalized by the number of the removed values. The model with the lowest total validation score by the final training epoch is kept as the resulting trained model.

\subsection{Wildfire observation model: multivariate zero-modified probability density}\label{sec::WOM-ZMPD}

A cursory data analysis shows that marginal distributions of BA and CNT features each appear to be bimodal, as a single bell-like distribution on the log-scale which is mixed with a positive probability mass at 0. Moreover, at each spatiotemporal location where BA and CNT are both observed, ie.\ not missing, they are either both zero or both positive real numbers. It seems reasonable to assume that if wildfire count is 0, burnt area must also be equal to 0, and vice versa. With sufficient generality, we define our wildfire observation model to be a product of probability distributions for each grid cell $x_{wh}$ of a data sample $\mathbf{x}$:
\begin{equation}
\mathbf{\varPhi}(\mathbf{x}\vert\mathbf{\boldsymbol{\eta}}) = \prod_{w=1}^W\prod_{h=1}^H\mathbf{\varPhi}_1(x_{wh}\vert\boldsymbol{\eta}_{wh}).
\end{equation}
$\mathbf{\varPhi}_1(x_{wh}\vert\boldsymbol{\eta}_{wh})$ is some zero-modified probability distribution of the pair $(\mathrm{CNT}, \mathrm{BA})_{wh}\in\mathbb{R}_{\geq 0}^2$ at the single spatial cell $(w,h)$:
\begin{equation}
\mathrm{\varPhi}_1\left(\mathrm{CNT}, \mathrm{BA}\vert p_{0,wh},\boldsymbol{\eta}^\prime_{wh}\right) = 
\begin{cases}
      p_{0,wh} & \text{for $\mathrm{CNT}=\mathrm{BA}=0$,}\\
      (1-p_{0,wh})f_\textrm{pos}\left(\mathrm{CNT}, \mathrm{BA}\vert \boldsymbol{\eta}^\prime_{wh}\right) & \text{for $\mathrm{CNT}>0$, $\mathrm{BA}>0$,}\\
      0 & \text{otherwise.}
    \end{cases}  
\end{equation}
where $p_{0,wh}$ is the probability mass at $\mathrm{CNT} = \mathrm{BA} = 0$ for grid cell $(w,h)$, and $f_\textrm{pos}\left(\cdot\vert\boldsymbol{\eta}^\prime_{wh}\right)$ is the probability density of the pair $(\mathrm{CNT}>0, \mathrm{BA}>0)$ parameterized by $\boldsymbol{\eta}^\prime_{wh}$. This probability distribution can be regarded as a Bernoulli coin flip between no wildfire at all and some bivariate distribution in the case of wildfires, at each separate grid cell.

We can further assume that the non-zero, strictly positive values of BA and CNT must have a single, well-localized mode on the logarithmic scale. A suitable choice of $f_\textrm{pos}$ is the bivariate log-normal probability density function $f_\textrm{ln}\left(\mathrm{CNT}, \mathrm{BA}\vert \boldsymbol{\mu}_{wh},\boldsymbol{\sigma}_{wh}\right)$ parameterized by $\boldsymbol{\mu}_{wh}\in\mathbb{R}^2$ and diagonal $\boldsymbol{\sigma}_{wh}\in\mathbb{R}^2$. Hence, to fully specify the zero-modified log-normal (ZMLN) observation model, the decoder needs to output $(1+2+2)\times W \times H$ parameters. Note that BA and CNT, if positive, are modelled as uncorrelated, but the decoder may implicitly learn and represent a correlation between their distribution parameters.

The single-mode assumption is quite restrictive and can be fully lifted. In this case we take $f_\textrm{pos}\left(\mathrm{CNT}, \mathrm{BA}\vert \boldsymbol{\eta}^\prime_{wh}\right)$ to be a product of two binned distributions for non-zero, positive CNT and BA. The bin edges are defined by the scoring scheme given by Eq.\ \ref{eq:score}, and the parameters $\boldsymbol{\eta}^\prime_{wh}$ are the probability masses for each bin at each grid cell ($w, h$). In total, the decoder for a zero-modified binned (ZMB) distribution needs to have the capacity to learn to output $(1+28+28)\times W \times H$ parameters, so the additional flexibility comes at a sizable computational cost.

\subsection{Convolutional neural network architecture of wildfire CMIWAE}
It is instructional to first describe the neural networks in a conventional MIWAE model and then build on it towards the architecture of the CMIWAE model used in this competition. MIWAE typically contains only two fittable neural networks: the encoder $g_{\boldsymbol{\gamma}}$ which serves to define the parameters of the variational distribution, and the decoder $f_{\boldsymbol{\theta}}$ the output of which defines the parameters of the observation model. Their architecture is chosen in such a way to be suitable for the data at hand.

The MIWAE encoder for image-like data such as the wildfire dataset would typically be realised as a deep convolutional neural networks and the corresponding decoder as a deep transposed convolutional neural network.\cite{Mattei19} These networks are a composition of $N_\textrm{lay}$ functions, so-called layers, where each layer is a composition of an affine mapping, a Batch Normalization layer \cite{Ioffe2015}, and an element-wise nonlinear function, so-called activation function. In order to best utilize expected spatial correlations in our wildfire data, for the affine mappings of the encoder we use the matrix convolutional operation \cite{Krizhevsky2012, Schlemper2017_2} in the two spatial dimensions. Similarly, for decoder affine maps, we use the matrix transposed convolutional operation. Batch Normalization layer facilitates better convergence during model training and generalization to unseen data. For the activation function we use the Softplus function given by $S(x):=\ln(1+\exp(x))$. The domain of the first layer of the encoder network is a space of three dimensional real tensors, having shape $d_\textrm{ch,0}\times W\times H$, where $d_\textrm{ch,0}$ signifies the number of features per grid cell at the input, so-called channels.

The codomain of the first layer that corresponds to the subsequent domain of the second encoder layer is a space of real tensors of shape $d_{\textrm{ch},1}\times W\times H$, where $d_{\textrm{ch},i}$ is the number of channels at the output of the $i$-th layer. Subsequent $N_\textrm{lay}-2$ encoder layers map real tensors shaped $d_{\textrm{ch},i-1}\times(W/2^{i-2})\times(H/2^{i-2})$ to real tensors shaped $d_{\textrm{ch},i}\times(W/2^{i-1})\times(H/2^{i-1})$, where $i=2,\ldots,N_\textrm{lay}-1$. The last encoder layer is a fully connected neural network consisting of a single affine map. The domain of the last layer is a space of real tensors shaped $d_{\textrm{ch},N_\textrm{lay}-1}\times(W/2^{N_\textrm{lay}-2})\times(H/2^{N_\textrm{lay}-2})$ and its codomain is a space of real matrices shaped $d\times 2$, where $d$ is the size of the latent space.

The matrix output of the encoder is finally interpreted as two $d$-dimensional real vectors representing parameters $\boldsymbol{\gamma}$ of the normal variational distribution $q_{\boldsymbol{\gamma}}$. The first is a vector of means, and the second one, after an additional element-wise Softplus transformation to guarantee positive values, is the vector of standard deviations. The decoder follows the encoder architecture in reverse, as input it takes a vector from the latent space of dimension $d$ and in its layers uses transposed convolutions instead of convolutions. The decoder outputs the parameters $\boldsymbol{\theta}$ of the observation model, the distribution $p_{\boldsymbol{\theta}}$.

\begin{figure}
\includegraphics[clip,width=1.0\linewidth]{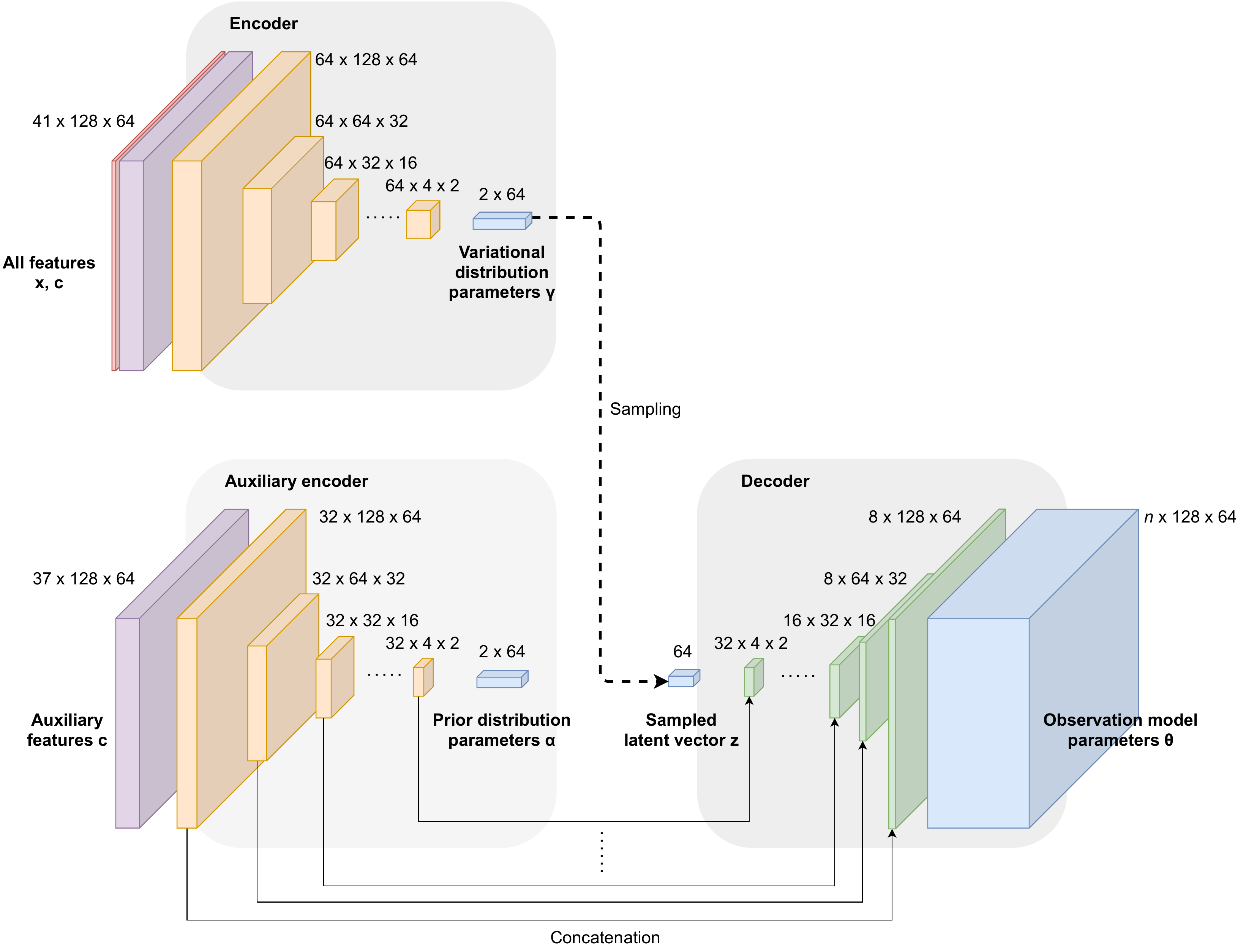}
\caption{Neural network architecture of the CMIWAE convolutional model used for prediction of missing wildfire data. Colored blocks represent the tensors flowing through the network layers. Three networks are used, the encoder, auxiliary encoder, and the decoder. The decoder structure is reversed with respect to the two encoders. Full-line arrows denote U-Net-like skip connections where a certain amount of channels from the intermediate results of the auxiliary encoder layers are passed to the corresponding decoder layers to be concatenated and used as input for the next decoder layer. Dashed arrow denotes sampling from the variational distribution. All three neural networks at their outputs calculate parameters for the corresponding conditional distributions. The shape of the parameters $\boldsymbol{\theta}$ depends on the specifics of the observation model. See text for more details.}
\label{fig:cmiwae}
\end{figure}

Expanding the model to CMIWAE, the flow of information becomes more complex and requires certain modifications to the neural architecture of the model as depicted by Fig.\ \ref{fig:cmiwae}. The auxiliary data $\mathbf{c}$ now needs to condition the prior, posterior, and variational distribution. The CMIWAE encoder becomes a function of both the data that might be missing and the auxiliary features, $g_{\boldsymbol{\gamma}}(\mathbf{x},\mathbf{c})$. We choose to concatenate the image-like tensors $\mathbf{x}$ and $\mathbf{c}$ along the spatiotemporal feature dimension and make the domain of the encoder be tensors with larger shape $d_\textrm{ch,0}\times W\times H$, $d_\mathrm{ch,0}=d_\mathrm{in}=41$.

Next, as mentioned in Sec.\ \ref{sec:CMIWAE}, the CMIWAE prior is parameterized by the auxiliary encoder $h_{\boldsymbol{\alpha}}(\mathbf{c})$. In our implementation, it follows the same architecture as the encoder except that $d^\prime_\mathrm{in,i}$ may differ and $d^\prime_\mathrm{in,0}=d_\mathrm{in}-4=37$, ie.\ the first layer receives as input auxiliary features only (the same features as the encoder except BA, CNT, and their two $\log$-transforms which might all be missing).

Finally, we need to devise an architecture for the CMIWAE decoder $f_{\boldsymbol{\theta}}(\mathbf{z},\mathbf{c})$. One could perhaps first map both $\mathbf{z}$ and $\mathbf{c}$ to the same shape using fully connected neural layers and then concatenate the results to form an input to further decoder layers. If the latent vector is expanded in this way to be concatenated with the auxiliary tensor, the large number of fittable parameters in the required fully connected layer presents a computational bottleneck. On the other hand, if the auxiliary data is compressed and concatenated with the latent vector, valuable information is lost already at the very first decoder layer.

We alleviate the issue of $\mathbf{z}$, $\mathbf{c}$ dimensionality by noting that a conventional MIWAE decoder's layers start with a latent vector and successively map it to larger image-like tensors. The shape of decoder's intermediary tensors, apart maybe the number of channels $d_{\textrm{ch},i}$, match in reverse the shapes of intermediate image-like tensors produced by the layers of the auxiliary encoder. Therefore, we can reuse the intermediary information from the auxiliary encoder and pass all or a part of its channels to the decoder, very much like the skip connections of the U-Net architecture.\cite{Shelhamer17} We have implemented the auxiliary encoder to output all of the intermediate tensors after every $i$-th layer, for every $i=1,\ldots,N_\textrm{lay}-1$. These outputs are then fed into the decoder at its matching layer by concatenating with intermediate tensors flowing through the decoder along the feature axis. As evidenced by our results (see Section \ref{sec:results}), this passthrough of information from auxiliary encoder directly to the encoder via skip connections is an important feature of our CMIWAE architecture, as it frees the model from needing to encode all of the always-complete auxiliary data $\mathbf{c}$ in the latent space.

\subsection{Model implementation and hyperparameters}\label{sec:nn_hyperparameters}

The CMIWAE model for USA wildfire missing data prediction is implemented in Python using the PyTorch library for GPU-accelerated computing \cite{Paszke2019}. The encoders and decoder are designed using $N_\textrm{lay}=7$ layers, where the number of encoder channels are $d_{\textrm{ch},i}=64$ for $i=1,\ldots,6$ and the number of auxiliary encoder channels are $d'_{ch,i}=32$ for $i=1,\ldots,6$. So there are $6$ layers of convolutions and a single fully connected layer. The number of decoder channels per layer of transposed convolutions are $32, 32, 16, 16, 8, 8$, and at every level the same number of channels are taken from the corresponding layer of the auxiliary encoder and concatenated. Notice, as auxiliary encoder has $32$ channels at every layer output, not all of these channels are used as decoder input due to performance reasons.

Every convolution or transposed convolution layer has a hyperparameter called the kernel size. For every such layer we use the kernel size of $5\times 5$. Further, a very important hyperparameter in the design of any generative latent variable model is the dimension $d$ of the latent space. We used the value $d=64$. Lastly, a hyperparameter specific to MIWAE-type models is the number of samples $K$ in the latent space taken from the prior. In theory, see Section \ref{sec:MIWAE}, it is best to take the largest value of $K$ allowed by hardware limitations. In our case, training a model with the value $K=768$ used up almost all of the available memory.

Additional so-called dropout layer is included as the first layer before every other layer in both encoders and the decoder. The function of this layer is to reduce overfitting during training \cite{Nitish2014} at the expense of introducing additional stochasticity by randomly zeroing out information passing through this layer, depending on a single hyperparameter $p\in[0,1]$, which is called the dropout percentage. More precisely, the dropout layer is an identity map on space $\mathbb{R}^k$ that is multiplied element-wise by a random vector $\mathbf{v}\in\{0,1\}^k$. Coordinates of $\mathbf{v}$ are independently sampled every time the dropout layer is evaluated. Each coordinate being $1$ with probability $1-p$ and $0$ with probability $p$. We used $p=10\%$ for every dropout layer throughout our model.

The total number of fittable parameters in our model is close to $10^6$.

\subsection{Optimizer hyperparameters}\label{sec:opt_hyperparameters}

Our models are trained using the fast.ai library \cite{Howard2018}. For the optimizer we employ the usual choice of Adam algorithm \cite{kingma2017adam}. One-cycle policy with cosine annealing \cite{smith2018superconvergence} is used for learning rate and momentum scheduling. This ensures better convergence of model parameters to a broad optimum that allows better generalization of the trained model \cite{smith2018disciplined}. We have used the following values of the optimizer hyperparameters: starting learning rate of $1.2\cdot 10^{-4}$, maximum learning rate of $3\cdot 10^{-3}$, final learning rate of $3\cdot 10^{-5}$, number of training epochs equal to $100$, and batch size equal to $12$. For Adam-specific hyperparameters, we have used: $\beta_2=0.99$, $\epsilon=10^{-5}$, and weight decay of $0.01$.

\subsection{Final prediction ensembling}\label{sec:ensembling}

Choosing only a single train-validation split for our model training and prediction resulted in high variation of validation scores due to the small size of the available dataset and poor representability. To counteract this, we use model ensembling to generate final predictions by pooling together predictions of many models trained with different train-validation splits. All the predicted distributions are weighted according to the likelihood of the observed features and mixed. This improved the validation score of the prediction significantly. Further, after the complete dataset was made available, we have confirmed the favorable effect of ensembling on final prediction scores, see Section \ref{sec:results}.

\subsection{Hardware requirements}

The main limitation in running our model using specified hyperparameters is the amount of GPU RAM needed to train the model. We have used a server equipped with 6 Nvidia Quadro RTX 5000 GPUs, each with 16\,GB of RAM. The training of a single model was distributed over all 6 GPUs and used up most of the total 96\,GB of GPU RAM. On the other hand, the required amount of CPU RAM used was moderate and peaked below 20\, GB. One full ensemble of models is trained and a single prediction evaluated in around 30 hours.

The amount of GPU RAM needed to train our model could be reduced by fine-tuning the model hyperparameters, perhaps significantly and without sacrificing the model performance. We however did not not search for optimal hyperparameters due to time requirements of repeated training of ensembles.

\section{Results and discussion}\label{sec:results}

Here we discuss the performance of our model that was measured in a series of tests, after the competition was finished and after the organiser supplied to us the full data, so that we can easily evaluate score by ourselves.

\begin{figure}[bth]
\includegraphics[clip,width=1.0\linewidth]{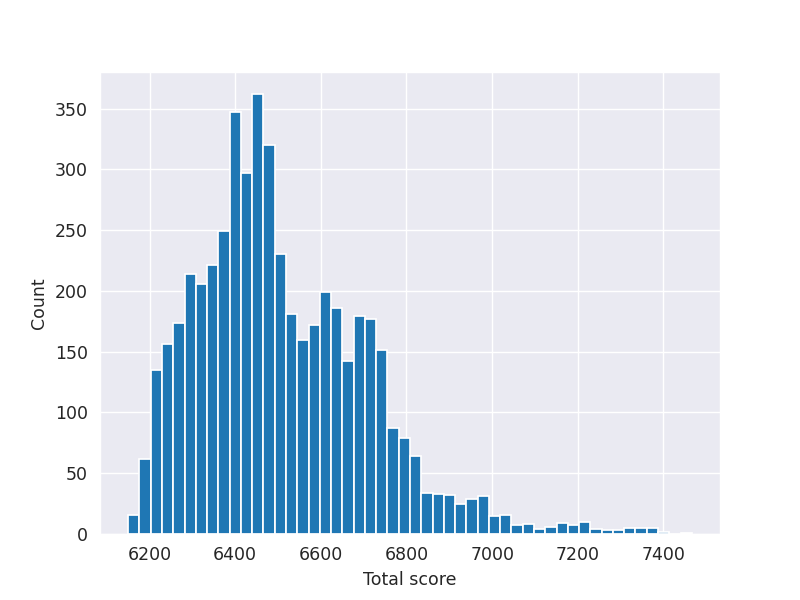}
\caption{Histogram of total scores calculated on predictions by single models. For each of $253$ trained models, $20$ predictions were evaluated, for a total of $5060$ predictions. The mean score of all predictions is $6511$, with a standard deviation of $200$. The large span of obtained total scores can be mitigated by model ensembling, see text.}
\label{fig:hist_scores}
\end{figure}

First we tried training single models using a random train-validation split out of possible 253 split combinations. We noticed a high variation in acquired validation score. We illustrate this effect on the total score calculated on complete data, see Figure \ref{fig:hist_scores}. The most likely cause of such a high variation is the nonrepresentativeness of each single train-validation split in a dataset with only 161 total monthly samples. To counter this, in the rest of the work we use model ensembling of 253 models as described in Section \ref{sec:ensembling}, using every possible train-validation split where validation is performed on two calendar years.

To estimate the influence of posterior family and auxiliary encoder on the prediction scores $S_\mathrm{CNT}$, $S_\mathrm{BA}$, and $S_\textrm{total}$, we trained several variants of CMIWAE ensembles with hyperparameters described in Sections \ref{sec:nn_hyperparameters} and \ref{sec:opt_hyperparameters}. The cumulative distribution of the missing data is recovered by sampling from the latent space of every model in the ensemble for $100$ times, to a cumulative of total 25,300 samples, as described in Section \ref{sec:cum_distro}.

The first variant employs the heavy-tailed zero-modified log-normal for the observation model family, as described in Section \ref{sec::WOM-ZMPD}. The second variant is a modified model with a disabled auxiliary encoder and a standard normal prior, effectively removing the parameters $\boldsymbol{\alpha}$ from the model. This variant may be considered a MIWAE-like model with auxiliary data present at encoder input. In the third variant, the skip connections from auxiliary encoder to the decoder are disabled, but the prior $p_{\boldsymbol{\alpha}}$ is learnable. In the fourth variant the prior is a fixed standard normal distribution, but the skip connections from auxiliary encoder to the decoder are kept. The last variant uses the zero-modified binned observation model. Due to hardware restrictions, the largest $K$ the zero-modified binned distribution was trained with was 300. For each of the ensemble variants, we evaluated $100$ different ensemble predictions, computed their scores and presented statistics on mean and standard deviation of scores, see Table \ref{table:scores}.

\begin{table}
\begin{tabular}{|cccc|ccc|}
\hline
\multicolumn{4}{|c|}{model ensemble variant}                                                                                                                                                                                                                                                                                           & \multicolumn{3}{c|}{scores (mean $\pm$ std)}                                                      \\ \hline
\multicolumn{1}{|c|}{\begin{tabular}[c]{@{}c@{}}obser-\\ vation\\ model\end{tabular}} & \multicolumn{1}{c|}{\begin{tabular}[c]{@{}c@{}}fittable\\ $p_{\boldsymbol{\alpha}}(\mathbf{z}\vert\mathbf{c}_{i})$\end{tabular}} & \multicolumn{1}{c|}{\begin{tabular}[c]{@{}c@{}}skip connections\\ from aux.\ encoder\\ to decoder\end{tabular}} & $K$ & \multicolumn{1}{c|}{$S_\mathrm{CNT}$} & \multicolumn{1}{c|}{$S_\mathrm{BA}$} & $S_\textrm{total}$ \\ \hline
\multicolumn{1}{|c|}{ZMLN}                                                            & \multicolumn{1}{c|}{yes}                                                                                                      & \multicolumn{1}{c|}{yes}                                                                                 & 768 & \multicolumn{1}{c|}{$2846\pm 18$}     & \multicolumn{1}{c|}{$3315\pm 7$}     & $6161\pm 20$       \\ \hline
\multicolumn{1}{|c|}{ZMLN}                                                            & \multicolumn{1}{c|}{no}                                                                                                       & \multicolumn{1}{c|}{no}                                                                                  & 768 & \multicolumn{1}{c|}{$2978\pm 17$}     & \multicolumn{1}{c|}{$3425\pm 8$}     & $6403\pm 19$       \\ \hline
\multicolumn{1}{|c|}{ZMLN}                                                            & \multicolumn{1}{c|}{yes}                                                                                                      & \multicolumn{1}{c|}{no}                                                                                  & 768 & \multicolumn{1}{c|}{$2955\pm 19$}     & \multicolumn{1}{c|}{$3402\pm 8$}     & $6357\pm 21$       \\ \hline
\multicolumn{1}{|c|}{ZMLN}                                                            & \multicolumn{1}{c|}{no}                                                                                                       & \multicolumn{1}{c|}{yes}                                                                                 & 768 & \multicolumn{1}{c|}{$2845\pm 17$}     & \multicolumn{1}{c|}{$\mathbf{3304\pm 5}$}     & $6149\pm 18$       \\ \hline
\multicolumn{1}{|c|}{ZMB}                                                             & \multicolumn{1}{c|}{yes}                                                                                                      & \multicolumn{1}{c|}{yes}                                                                                 & 300 & \multicolumn{1}{c|}{$\mathbf{2782\pm 14}$}     & \multicolumn{1}{c|}{$3339\pm 7$}     & $\mathbf{6121\pm 18}$       \\ \hline
\end{tabular}
\medskip
\caption{Scores for the reconstructed wildfire count CNT and size BA obtained by different model ensemble variants, smaller value is better. Bold numbers highlight the model ensemble variant with the best score.}
\label{table:scores}
\end{table}

Indeed, compared with Fig.\ \ref{fig:hist_scores}, model ensembling decreases the variation in score significantly and to similar standard deviations independent of the variant of ensemble. The best performing variant (lowest scores) uses the ZMB observation model which we did not originally use for competition. This is a surprising result because the fittable binned distribution has the most flexible shape that still makes sense to use for the problem at hand, ie.\ it carries the least amount of expert knowledge. Moreover, it is trained with a small $K$ which should provide a disadvantage according to Eq.\  \ref{eq:bound_limit}.

The worst performance (highest scores) is obtained by the two variants not using skip connections from the auxiliary encoder to the decoder. This result underlies the architectural importance of skip connections for problems on a spatial grid with lots of helpful auxiliary correlates. As expected, out of the two worst variants, the better one is using learnable prior $p_{\boldsymbol{\alpha}}$, which allows the auxiliary features to help improve the score somewhat, while the worst one is using the fixed, standard normal distribution prior.

Interestingly, it is difficult to say whether the same relation holds in the case of the other two ZMLN variants that do employ the skip connections. There, the learnable prior appears to slightly worsen the final score, but the small difference is well within the standard deviation and can be attributed to stochasticity in the process of model training. It is unclear whether a fittable prior generally does not improve model performance compared to a fixed one, or if this occurrence is specific to our hyperparameters (or maybe, but less likely so, the wildfire dataset itself). As a most interesting simplification, if a dataset consists of vector-like, easy-to-concatenate quantities and a fixed, non-parameterized prior is preferred, one can streamline CMIWAE by replacing the auxiliary encoder with direct concatenation of $\mathbf{c}$ to inputs of both the (fully-connected) encoder and decoder, and training a MIWAE-like model with the log-likelihood objective marginalized over all auxiliary features.

The thick-tailed zero-modified log-norm observation model, used for each grid cell, fared slightly worse than the zero-modified binned distribution. While wildfire counts and sizes may be modeled by thick-tailed marginal distributions, our results raise the question whether their Bayesian probabilities conditioned on all the other observed features also need to be considered heavy-tailed. Unfortunately, the resolution of CNT and BA bins employed in this work is too low to reliably conclude whether their ZMB tails are thick. Here, a higher resolution for large values of CNT and BA would be preferable, but likely unfeasible to compute. It may be fruitful to explore other probability distributions with different parameterizable tails, but such work is out of scope of this paper.

Considering the very good results of CMIWAE in tackling the problem of wildfire missing data, it would be advantageous to learn which features contribute most to the end prediction. The image-like nature of wildfire data motivated the pervasive use of convolutional neural networks (CNNs) in this work. One disadvantage of CNN architectures is their poor interpretability. There is much ongoing work in interpreting classifiers, specifically CNN image-to-class models, and estimating the spatially-distributed feature importance, so-called saliency, for the final classification, see eg.\ \cite{Simonyan14,Simonyan15,Kindermans19,Tomsett20}. To the best of our knowledge there are no such techniques for image-to-image CNN models, such as the VAE/IWAE family, as each feature on each grid cell of the input may influence every one of the many grid cells at the output. One might build a saliency map, for each of the data samples and each observation model parameter at some specific grid cell at the output, but these maps would necessarily be many-dimensional and data-dependent. Alternatively, in a future labor-intensive work one might train models on restricted sets of features and assess the impact of removal of certain input channels on some model performance metric (log-likelihood or scores).

In the end, we wish to note that our competition-winning prediction achieved a slightly better score than the ones discussed above. It was obtained by an ensemble with the exact architecture of the first variant above (CMIWAE with zero-modified log-normal distribution) but a somewhat modified training procedure. First, smaller additional validation masks were used, ie.\ less data was withheld for the validation score computation. Second, a larger amount of time-costly sampling was used to recover the cumulative distribution of the missing data, as much as $1024$ samples from each model. Third, only $112$ models were ensembled trained on randomly selected train-validation splits. Seeing that the methodology of our predictions submitted to the competition was insufficiently rigorous, we decided to improve on it for this paper. The somewhat diminished (but still winning) score of presented models mainly suggest that the validation score masks we used for this paper may be overly conservative.

\section{Conclusion}

In this work we present a deep latent variable generative model based on the missing importance-weighted autoencoder \cite{Mattei19} which is used to recover conditional distributions of missing data. The CMIWAE model is constructed to accomodate the large number of auxiliary features guaranteed to be always present and trained to maximize log-likelihood of the observed features of interest. We successfully apply it to the problem of recovering missing wildfire size and frequency distributions from the image-like USA wildfire dataset \cite{Opitz22}, with little to no feature engineering, minimal required domain knowledge, and an end-to-end training procedure from data to final probability distribution prediction. The introduction of a second, auxiliary encoder which is connected directly to the decoder network \cite{Shelhamer17} allows the auxiliary information to greatly improve the prediction results. The presented model architecture can immediately be applied to recover missing image-like or tensor-like data from other domains which are conditioned on classes, tags, or other images. Further work is necessary to examine model interpretability and feature importance, the influence of the choice of observation model distribution on prediction performance, and perhaps even test if the approach generalizes to the case of data missing-not-at-random (MNAR)\cite{Mattei19}.

\backmatter

\bmhead{Acknowledgments}
We thank Stjepan \v{S}ebek and Josip \v{Z}ubrini\'{c} for valuable discussions and help in data preparation. This research was supported by: Croatian Science Foundation (HRZZ) grant PZS-2019-02-3055 from ``Research Cooperability'' program funded by the European Social Fund.

\bibliography{sn-bibliography}


\end{document}